\documentclass[letter, 10 pt, conference]{ieeeconf}
\IEEEoverridecommandlockouts                              
\usepackage{times}
\usepackage{epsfig}
\usepackage{graphicx}
\usepackage{amsmath}
\usepackage{amssymb}
\usepackage{multirow}

\usepackage{graphicx}
\usepackage{epsfig}
\usepackage{epic,eepic}
\usepackage{times}
\usepackage{mathptmx}
\usepackage{amsfonts}
\usepackage{amsmath}
\usepackage{cite}
\usepackage{color}

\usepackage{times}

\definecolor{red}{rgb}{1,0,0}
\definecolor{green}{rgb}{0,1,0}
\definecolor{blue}{rgb}{0,0,1}
\definecolor{violet}{rgb}{1,0,1}
\definecolor{cyan}{cmyk}{1,0,0,0}
\definecolor{magenta}{cmyk}{0,1,0,0}
\definecolor{yellow}{cmyk}{0,0,1,0}

\definecolor{white}{rgb}{1,1,1}

\usepackage{jumoline}

\newcommand{\CO}[2]{}

\newcommand{\CommentOut}[1]{}

\begin{document}

\newcommand{\FIG}[3]{
\begin{minipage}[b]{#1cm}
\begin{center}
\includegraphics[width=#1cm]{#2}\\
{\scriptsize #3}
\end{center}
\end{minipage}
}

\newcommand{\FIGU}[3]{
\begin{minipage}[b]{#1cm}
\begin{center}
\includegraphics[width=#1cm,angle=180]{#2}\\
{\scriptsize #3}
\end{center}
\end{minipage}
}

\newcommand{\FIGm}[3]{
\begin{minipage}[b]{#1cm}
\begin{center}
\includegraphics[width=#1cm]{#2}\\
{\scriptsize #3}
\end{center}
\end{minipage}
}

\newcommand{\FIGR}[3]{
\begin{minipage}[b]{#1cm}
\begin{center}
\includegraphics[angle=-90,clip,width=#1cm]{#2}
\\
{\scriptsize #3}
\vspace*{1mm}
\end{center}
\end{minipage}
}

\newcommand{\FIGRpng}[5]{
\begin{minipage}[b]{#1cm}
\begin{center}
\includegraphics[bb=0 0 #4 #5, angle=-90,clip,width=#1cm]{#2}\vspace*{1mm}
\\
{\scriptsize #3}
\vspace*{1mm}
\end{center}
\end{minipage}
}

\newcommand{\FIGpng}[5]{
\begin{minipage}[b]{#1cm}
\begin{center}
\includegraphics[bb=0 0 #4 #5, clip, width=#1cm]{#2}\vspace*{-1mm}\\
{\scriptsize #3}
\vspace*{1mm}
\end{center}
\end{minipage}
}

\newcommand{\FIGtpng}[5]{
\begin{minipage}[t]{#1cm}
\begin{center}
\includegraphics[bb=0 0 #4 #5, clip,width=#1cm]{#2}\vspace*{1mm}
\\
{\scriptsize #3}
\vspace*{1mm}
\end{center}
\end{minipage}
}

\newcommand{\FIGRt}[3]{
\begin{minipage}[t]{#1cm}
\begin{center}
\includegraphics[angle=-90,clip,width=#1cm]{#2}\vspace*{1mm}
\\
{\scriptsize #3}
\vspace*{1mm}
\end{center}
\end{minipage}
}

\newcommand{\FIGRm}[3]{
\begin{minipage}[b]{#1cm}
\begin{center}
\includegraphics[angle=-90,clip,width=#1cm]{#2}\vspace*{0mm}
\\
{\scriptsize #3}
\vspace*{1mm}
\end{center}
\end{minipage}
}

\newcommand{\FIGC}[5]{
\begin{minipage}[b]{#1cm}
\begin{center}
\includegraphics[width=#2cm,height=#3cm]{#4}~$\Longrightarrow$\vspace*{0mm}
\\
{\scriptsize #5}
\vspace*{8mm}
\end{center}
\end{minipage}
}

\newcommand{\FIGf}[3]{
\begin{minipage}[b]{#1cm}
\begin{center}
\fbox{\includegraphics[width=#1cm]{#2}}\vspace*{0.5mm}\\
{\scriptsize #3}
\end{center}
\end{minipage}
}

% \renewcommand{\FIGpng}[5]{~} \renewcommand{\FIGR}[3]{~} \renewcommand{\FIG}[3]{~} \renewcommand{\FIGU}[3]{~}

% \newcommand{\noeditage}[1]{#1} \newcommand{\editage}[1]{}
% \onecolumn \renewcommand{\editage}[1]{#1} \renewcommand{\noeditage}[1]{}

\newcommand{\figA}{
\begin{figure}
\begin{center}
\FIG{10}{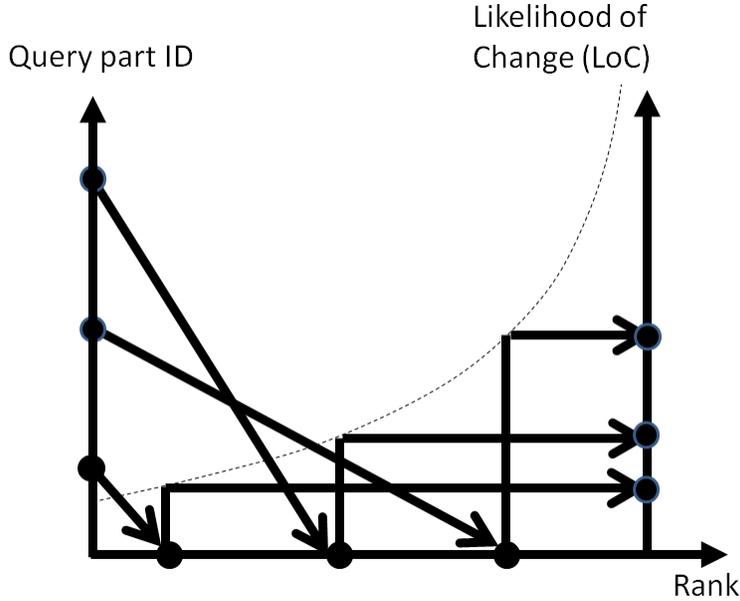}{}%
\SW{
\caption{%
In the detection-by-localization scheme, self-localization performance (i.e., rank value of the ``ground-truth" reference image in the map database) is used as a measure of likelihood-of-change (LoC) for change detection. The new generalized problem of object-level change detection, addressed in this paper, takes object-level subimages (instead of a full image) as query input for the self-localization, and outputs a subimage-level pixel-wise LoC map. 
}\label{fig:a}
}{
\caption{
}\label{fig:a}
}
\end{center}
\end{figure}
}

\newcommand{\figB}{
\begin{figure}
\begin{center}
\FIG{10}{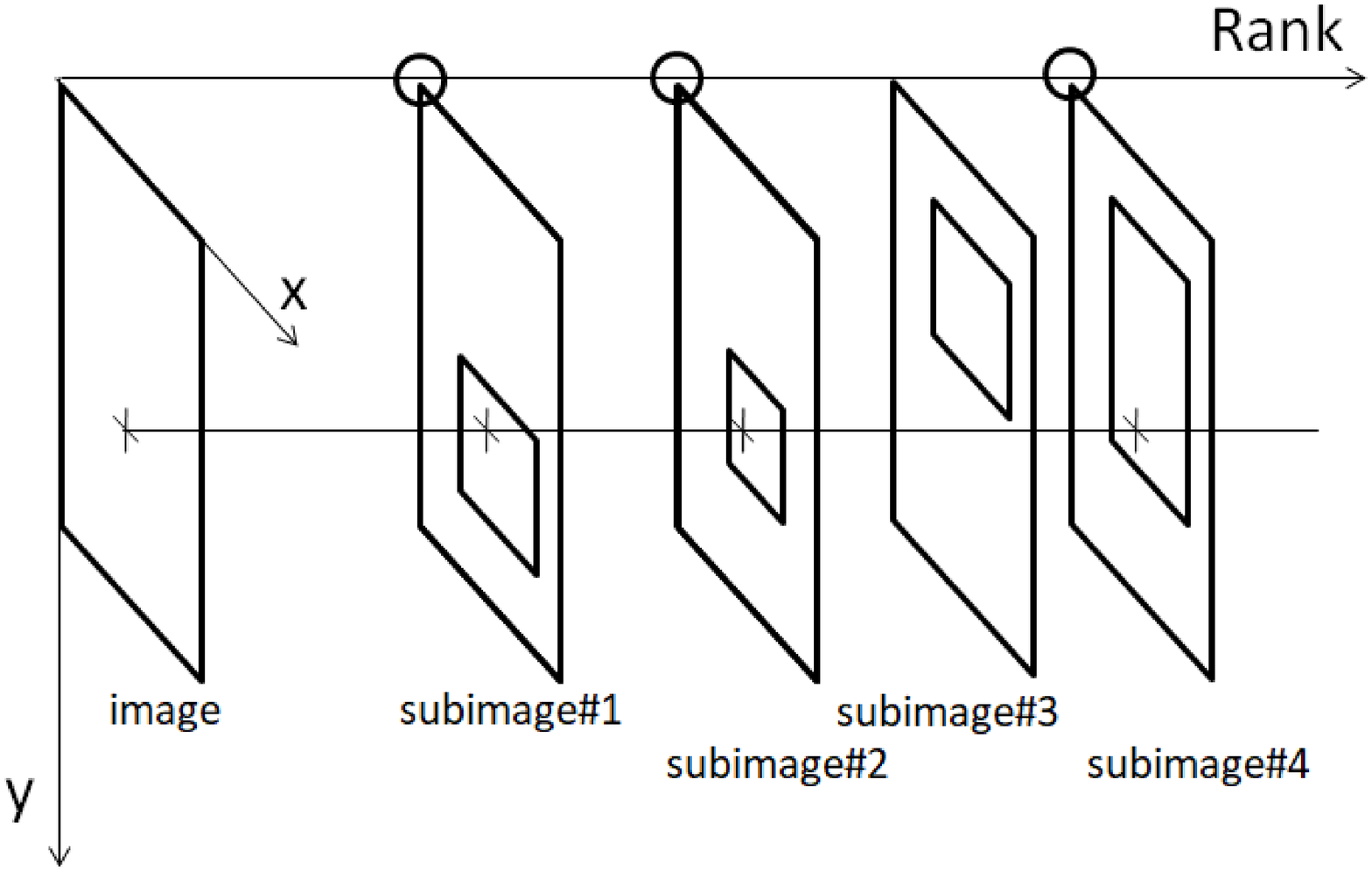}{}
\SW{
\caption{Estimating LoC map.
We consider the new setting where each pixel may belong to multiple subimages (referred to as qBBs).
Each subimage is then input to a self-localization system to obtain a rank value,
and then the rank value of each pixel is computed from these subimages which the pixel belongs to.
In the figure, a rectangle indicates the bounding box of an image or a subimage, and a '+' mark indicates the pixel of interest. 
}\label{fig:b}
}{
}
\vspace*{-3mm}
\end{center}
\end{figure}
}

\newcommand{\figC}{
\begin{figure}
\begin{center}
\FIG{10}{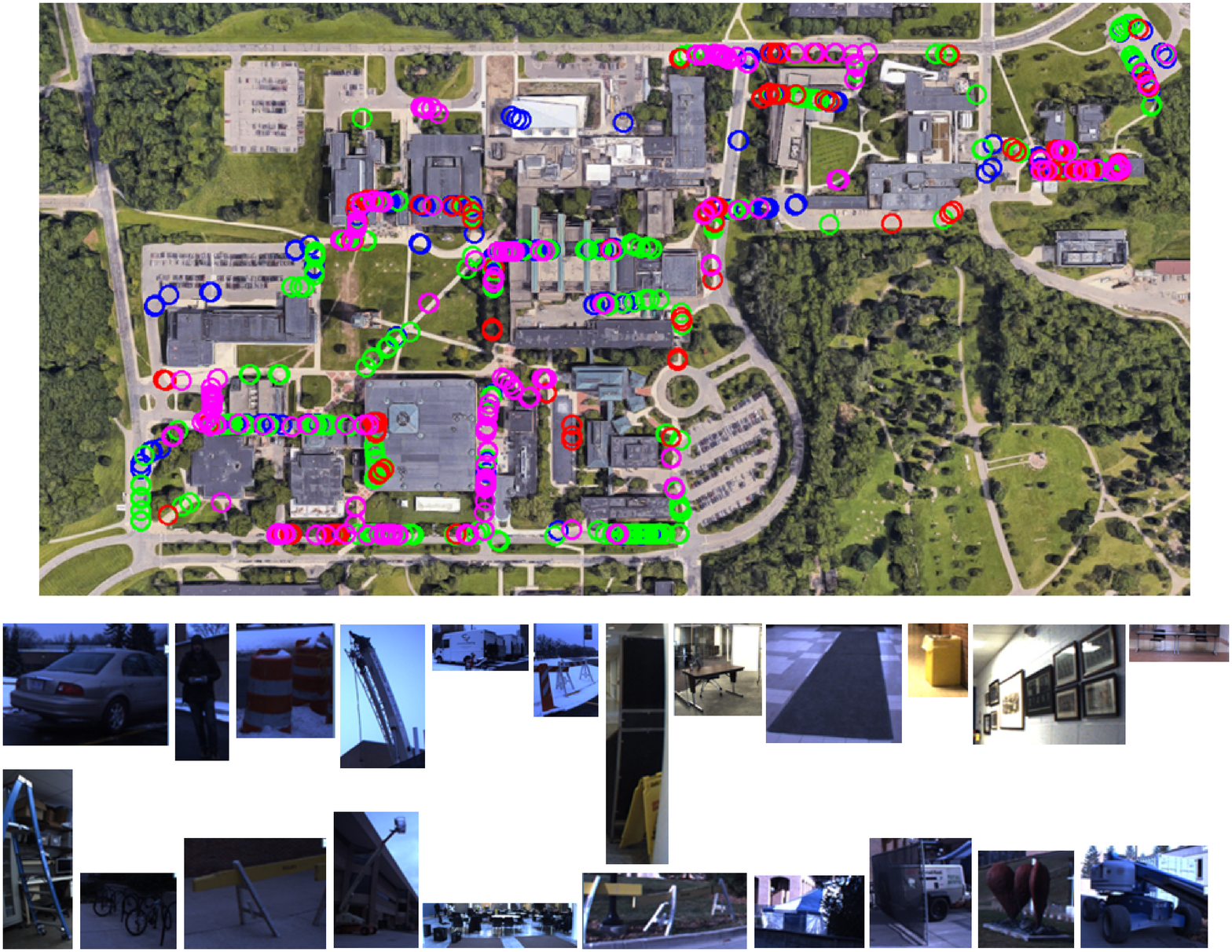}{}%
\SW{
\caption{Experimental settings.
Top: the workspace and the robot's viewpoints of the test images.
Bottom: examples of change objects.
}\label{fig:env}
}{
}
\end{center}
\end{figure}
}

\newcommand{\figD}{
\begin{figure}
\FIGpng{8}{cd18_fig4.png}{}{5168}{8380}
\SW{
\caption{
Examples of LoC maps.
From left to right,
each panel shows
an input image,
the LoC map by the proposed ``rank fusion" method,
the LoC map by the ``score sum",
and the LoC map by the ``score max".
}\label{fig:examples}
}{
}
\end{figure}
}

\newcommand{\vs}{\hspace*{-1mm}}

\newcommand{\tabA}[2]{
\begin{table}[t]
\begin{center}
\caption{Performance results}\label{tab:a}
\begin{tabular}{l|rrrrr|}
AP & \multicolumn{5}{c|}{$RoC_{max}^{ - }$} \\ 
&\vs 0.01 &\vs 0.02 &\vs 0.03 &\vs 0.04 &\vs 0.05 \\ \hline 
rank fusion &\vs 0.74 &\vs 0.73 &\vs 0.72 &\vs 0.72 &\vs 0.72 \\   
rank fusion ($\le$ 2) &\vs 0.72 &\vs 0.74 &\vs 0.73 &\vs 0.74 &\vs 0.72\\  
rank fusion ($\le$ 3) &\vs 0.72 &\vs 0.74 &\vs 0.73 &\vs 0.74 &\vs 0.74\\  
rank w/o fusion &\vs 0.69 &\vs 0.68 &\vs 0.68 &\vs 0.68 &\vs 0.69 \\   
score max &\vs 0.79 &\vs 0.77 &\vs 0.76 &\vs 0.76 &\vs 0.77 \\    
score sum &\vs 0.73 &\vs 0.71 &\vs 0.70 &\vs 0.70 &\vs 0.71 \\ \hline 
\end{tabular}
\end{center}
\end{table}
}
\newcommand{\SW}[2]{#1}

\renewcommand{\CO}[2]{#2}

\CO{}{\onecolumn }

\newcommand{\editage}[2]{#1}

\title{\LARGE \bf
Detection-by-Localization: Maintenance-Free Change Object Detector
}
\author{Tanaka Kanji 
\thanks{Our work has been supported in part by 
JSPS KAKENHI 
Grant-in-Aid 
for Scientific Research (C) 26330297, and (C) 17K00361.}
\thanks{K. Tanaka is with Faculty of Engineering, University of Fukui, Japan. 
{\tt\small tnkknj@u-fukui.ac.jp}}
\thanks{We would like to express our sincere gratitude to Takuma Sugimoto, Rino Ide and Kousuke Yamaguchi for development of deep learning architecture, and initial investigation on change detection tasks on the dataset, which helped us to focus on our Detection-by-Localization project.}}
\maketitle

\begin{abstract} 
\SW{\CO{
}{
Recent researches demonstrate that self-localization performance is a very useful measure of likelihood-of-change (LoC) for change detection. In this paper, this ``detection-by-localization" scheme is studied in a novel generalized task of object-level change detection. In our framework, a given query image is segmented into object-level subimages (termed ``scene parts"), which are then converted to subimage-level pixel-wise LoC maps via the detection-by-localization scheme. Our approach models a self-localization system as a ranking function, outputting a ranked list of reference images, without requiring relevance score. Thanks to this new setting, we can generalize our approach to a broad class of self-localization systems. Our ranking based self-localization model allows to fuse self-localization results from different modalities via an unsupervised rank fusion derived from a field of multi-modal information retrieval (MMR). This study is a first step towards a maintenance-free approach, minimizing maintenance cost of map maintenance system (e.g., background model, detector engine).
}}{
}

\end{abstract}

\section{Introduction}\label{sec:intro}

\SW{
To maintain an environment map in dynamic environments, a robotic visual SLAM system must be capable of detecting changed objects on the map. To be effective, a map maintenance capability must fulfil two requirements: the robot must know where it is (i.e., map-relative robot self-localization), and it must differentiate between changed objects and nuisances (i.e., robot-centric change detection). Motivated by the recent maturity of self-localization techniques (e.g., visual place recognition, loop-closure detection, map-matching), we propose the reuse of self-localization capability to address the latter requirement (i.e., change detection).
}{
}

\SW{
Recent studies demonstrated that self-localization performance is a very useful measure of the likelihood of change (LoC) for change detection. The experience-based mapping framework originally proposed in \cite{churchill2013experience}
is a good example. In that framework, an environment map was permanently augmented by multiple sub-maps (i.e., visual experiences) using data from differing environment conditions. If self-localization with such a map performs sufficiently well, it covers the encountered conditions and there is no need to update the map (i.e., low LoC). However, if self-localization performance is poor, the map does not cover the encountered conditions during the sortie (i.e., high LoC).
}{
}

\SW{
Such a new change detection scheme, referred to as ``detection-by-localization" in the current study, has two main advantages:
\begin{enumerate}
\item
No additional storage or detector engine (but only existing map database and localization engine) is required;
\item
The degration of map quality (i.e., need of map update) in terms of self-localization performance can be directly detected.
\end{enumerate}
}{
}

\SW{
This paper addresses a new detection-by-localization task that detects changes at subimage- or object- levels (Fig. \ref{fig:a}). Unlike previous methods that classify a given query image as ``change" or ``no-change", we not only classify a query image but also detect changed objects (in the form of bounding boxes) within the image frame.
The relationship between image- and object-level change detection is analogous to the relationship between image \cite{sivic2003video}
and object localization \cite{jiang2012randomized} in the computer vision literature.
Object-level detection becomes an even more challenging task than image-level detection, owning to the requirements of object segmentation and low image resolution. 
Importantly, object-level change detection is common in robotics 
and has many important applications, 
including
patrol robot in partially changing environments \cite{drews2010novelty} 
and
object segmentation via change detection \cite{finman2013toward}. 
However, object-level change detection has not been explored in the context of detection-by-localization.
}{
}

\SW{
In our task scenario, a ranked list of reference or background images retrieved from a map database by a self-localization system is viewed as the measure of self-localization performance. Formally, we consider a detection-by-localization scenario in which reliable viewpoint measurement (referred to as ``ground-truth" viewpoint) is available \cite{churchill2013experience}, and we measure self-localization performance (i.e., LoC) by the rank value assigned to the ``ground-truth" reference image. The self-localization model is inspired by our previous work on unsupervised part-based scene modeling, in which the popular bag-of-visual-features (BoVF) self-localization scheme \cite{sivic2003video} is extended in two ways. 
\begin{enumerate}
\item
Instead of point features (e.g., SIFT), discriminative subimage-level features (referred to as ``scene parts") are used to model a query/reference scene. 
\item
Reference subimages are retrieved using query subimages, and are aggregated into an image-level decision (i.e., ranked list). 
\end{enumerate}
In our contribution, 
\begin{itemize}
\item
we explain how such an image-level self-localization model can be adopted to the novel task of subimage-level change detection, 
\item
by addressing how a ranked list can be interpreted as a subimage-level pixel-wise LoC map, 
\item
and by exploring ways of aggregating subimage-level LoC maps into the final decision of a single image-level LoC map. 
\end{itemize}
Our unsupervised ranking based approach is an advantage, because it does not require training examples on the image collection.
This is very important, because we must deal with increasingly complex maps and images, and automatic collection of their ground-truth data by the robot-self is not a trivial task.
}{
}

\editage{
\figA
}{}

\SW{
Presently, most image change detection algorithms \cite{ccr1} focus on pairwise image comparison, to differentiate a given query-background-image-pair. It is straightforward to adopt such a method to object-level change detection by introducing an image segmentation preprocessing step \cite{im2008object}. This has recently achieved state-of-the-art performance  with a weakly-supervised setting \cite{aaai17cd}. However, these image-to-image differencing methods are  not directly applicable to real-time SLAM systems. First, they are too expensive (on average) to compute in real-time. Moreover, they require memorization and maintenance of many background images proportional to the map size. Some alternative popular approaches are those that classify a sole query image instead of a query-background-image-pair as a change or a no-change, compared to an offline pretrained background model. Such an approach has the potential to realize model compactness by using compressed background models such as BoVFs, geometric model and compact manifold learning. However, its storage overhead is proportional to map size and increases in unbounded fashion. Moreover, updating this background model significantly complicates the map updating process. To suppress such computational time and space complexities by reusing existing resources of map database and localization engine is an objective of the current study. 
}{
}

\SW{
}{
}

\SW{
}{
}

\section{Detection-by-Localization Framework}

\SW{
We consider a new detection-by-localization framework 
consisting of a self-localization system and a change detection system, in which self-localization performance 
(i.e., rank value of the ``ground-truth" reference image as explained in Section \ref{sec:intro})
is used as an LoC measure for change detection. Both self-localization and change detection can be confused by differing fields-of-view (FoV), occlusions, and seasonal changes of object appearance or appearing of visible objects.
}{
}

\subsection{Scene Model}\label{sec:sm}

\SW{
The scene model is assumed to be a BoVF scene model \cite{sivic2003video}. The vast majority of real-time visual SLAM systems (i.e., iBOW-LCD \cite{Garcia-Fidalgo2018}) employs BoVF models. 
Whereas other scene modeling schemes such as 
the deep ConvNet classifier have been studied in the self-localization literature. 
The BoVF scheme is advantageous to computational speed, compactness, and discriminativity, and has been a de facto standard method for real-time self-localization tasks, such as loop-closure detection. 
}{
}

\subsection{Self-Localization Model}

\SW{
The self-localization system is modeled as a ranking function.
That is,
it aims to output a ranked list of all reference subimages in descending order of similarity from a given query subimage.
If necessary, each query subimage is resized to an appropriate size before being input to the self-localization system.
This model is valid for most self-localization systems,
which are based on image retrieval formulations.
}{
}

\SW{
Not only are rank lists output by self-localization algorithms, but some relevance scores for individual reference images (e.g., TF-IDF, conditional probabilities, maximal consistencies) are also output. However, it is not straightforward to interpret such relevance scores to LoC values within a detection-by-localization framework. We reserve this issue for future work.
}{
}

\SW{
Image-level and subimage-level rank values are available as output of any self-localization system.
The BoVF-based self-localization systems first evaluate subimage-level similarities between each query subimage and each reference subimage,
and then aggregate their similarity values to obtain the image-level similarity scores or rank lists.
It is straightforward to output rank lists 
in the order of subimage-level similarity
for individual query subimages.
}{
}

\subsection{Change Detection Model}

\SW{
The change detection model follows the standard formulation of image change detection \cite{ccr1}. Given a query image, it aims to estimate a pixel-wise LoC map of the image, which is then converted to a pixel-wise binary change mask.
Its performance is evaluated in terms of 101-point (i.e., 0.00, 0.01, $\cdots$, 1.00) interpolated average precision (AP) \cite{yolo}.
}{
}

\editage{
\figB
}{}

\section{Object-Level Change Detection}

\SW{
This section provides detailed methods 
for object-level change detection
via detection-by-localization (Fig. \ref{fig:b}).
}{
}

\subsection{Scene Modeling}

\SW{
The scene modeling segments a given query/reference image to yield a pool of useful scene parts (in the form of bounding boxes) or subimages, so that the segmented subimages remain consistent between the reference and an unseen query images. This problem is referred to as ``consistent part segmentation" \cite{kanji2015unsupervised}. This problem is significantly ill-posed, owning to differing FoVs, occlusions, and seasonal changes of object appearance. Even state-of-the-art segmentation algorithms are far from perfect, producing several false positive parts. We address this issue by hypothesizing a relatively large number of subimages and verifying them in the spirit of majority voting \cite{jiang2012randomized}, as we explain in Section \ref{sec:dbl}.
}{
}

\SW{
}{
}

\SW{
We use both unsupervised and supervised segmentation techniques,
inspired by our previous work \cite{iv18sugimoto}.
Unsupervised segmentation techniques (e.g., BING \cite{cheng2014bing}) provide category-independent object proposals, even for objects with unseen classes.
We use a set of five pre-defined bounding boxes for every image.
For an image with width $w$ and height $h$, these five bounding boxes are defined as 
$[w/3, 2w/3]$$\times$$[h/3, 2h/3]$, 
$[0, 2w/3]$$\times$$[0, 2h/3]$, 
$[w/3, w]$$\times$$[0, 2h/3]$, 
$[0, 2w/3]$$\times$$[h/3, h]$, and $[w/3, w]$$\times$$[h/3, h]$.
Supervised segmentation techniques (e.g., YOLO \cite{yolo}) provide more precise object regions supported by rich semantic information for objects with known classes. We use YOLO with threshold value of 0.05. It should be noted that whereas supervised techniques are trained on pre-defined object classes, it is often useful to propose unseen objects, 
having a visually similar appearance to pre-defined objects.
Both supervised and unsupervised proposals are beneficial, 
because objects with both pre-defined and unseen classes can be changed objects.
}{
$[w/3, 2w/3]$$\times$$[h/3, 2h/3]$, 
$[0, 2w/3]$$\times$$[0, 2h/3]$, 
$[w/3, w]$$\times$$[0, 2h/3]$, 
$[0, 2w/3]$$\times$$[h/3, h]$, 
$[w/3, w]$$\times$$[h/3, h]$
}

\subsection{Self-Localization}

\SW{
In \cite{Merrill2018RSS}, a novel AE-based ConvNet for loop-closure
detection was presented. 
It was designed to be an unsupervised, convolutional AE
network architecture, tailored for loop-closure, and
amenable for efficient, robust place recognition.
Its performance was evaluated via
extensive comparison studies of the 
deep loop-closure model against state-of-the-art
methods on different datasets. 
Additionally, the histogram of oriented gradients (HoG) was employed to compress images while preserving salient features and projective transformations (i.e., homography). 
In contrast, 
we are interested in 
the basic effectiveness of an AE-based self-localization,
and we 
implement
the convolutional AE in a rather simplified setting without using 
a HoG -based extension.
Our AE consists of three convolutional layers and each convolutional kernel size is 3$\times$3,
employing max-pooling, batch-normalization and ReLU activation for each layer.
Each input subimage is resized to 256$\times$256,
and mapped by the AE to a 16,388-dim feature vector.
}{
max-pooling, batch-normalization 
256$\times$256
16388
}

\subsection{Detection-by-Localization}\label{sec:dbl}

\SW{
Following our previous research in \cite{kanji2015unsupervised}, we modeled individual scene parts belonging to different modalities, and adopted rank fusion techniques, derived from the field of multi modal information retrieval (MMR) \cite{atrey2010multimodal}. MMR techniques are originally designed to deal with increasingly complex document collections, corresponding to subimage or part collections in our application domain and queries, consisting of not only text modalities but also non-textual modalities such as visual words in image retrieval, geo-tags, user rate, etc. More formally, they aim to fuse multiple retrieval results and queries from different modalities into a single ranked list to make a final decision. Unsupervised approaches requiring no training data are desirable, as explained in Section \ref{sec:intro}. Such unsupervised MMR approaches are broadly classified into two categories: early fusion \cite{yan2006probabilistic} and late fusion \cite{cormack2009reciprocal}. 
Early fusion aims to fuse multiple queries from multiple modalities at the level of the input feature descriptor. This approach is advantageous in exploring correlation between multiple data modalities. However, it requires multiple queries to be converted to the same format prior to fusion. This is not applicable to a wide range of self-localization applications. 
Late fusion aims to fuse multiple queries at the output decision level (e.g., the level of relevance score or rank list). This approach is further divided into score fusion \cite{montague2002condorcet} and rank fusion \cite{cormack2009reciprocal}. 
Score fusions rely on the so-called raw-score-merging hypothesis and fuses relevance scores from multiple queries. However, as a key limitation, it requires each modality's retrieval to output a relevance score, significantly limiting its application.
However, rank fusion aims to fuse rank values from multiple queries. As a key advantage, this approach requires only ranked lists (commonly output by self-localization), and importantly it can outperform the score fusion approach in previous applications \cite{hsu2005comparing}.
}{
}

\SW{
Based on these considerations, we chose the rank fusion approach as our basis. In our previous study, the rank fusion approach was explored in a different context of image-level self-localization \cite{kanji2015unsupervised}. The current study
is based on this previous method, but with key differences: 
\begin{enumerate}
\item
The previous study aimed at image-level ranking, whereas the current study aims to obtain subimage-level pixel-wise rank values.
\item
The previous method took as input non-overlapping query subimages (from color-based segmentation), whereas the current study takes relatively large numbers of overlapping query subimages (from unsupervised/supervised object proposals). 
\end{enumerate}
To address this issue, we must deal with a novel task of pixel-wise rank fusion. 
Considering the new setting where each pixel may belong to multiple subimages (referred to as qBBs) as in Fig. \ref{fig:b}, we fuse them all to obtain the rank list for each pixel. Because the number of such qBB is different among different pixels, it is not straightforward to compare the fused rank list between them. Our previous ranking method in \cite{kanji2015unsupervised} assumed that the number of rank lists was pre-defined and fixed. Thus, it was not directly applicable to the current problem. We now address this issue by introducing a new parameter, $N(i)$, the number of times a document appear in the rank lists, as suggested in \cite{imhof2018study}. The modified rule for rank fusion takes the form:
\begin{equation}
R(i)=N(i) \times \sum_{k=1}^{N(i)} \frac{1}{R_k(i)}, 
\end{equation}
where $N(i)$ is the number of the $i$-th document appearing in the rank lists, and $R_k(i)$ is a rank value (of the ground-truth reference image) assigned by the $k$-th query's retrieval. If $N(i)$ is a constant, the function $R(i)$ reduces to the previous fusion rule. Our algorithm takes as input a collection of query bounding boxes (qBBs), and performs the following steps:
\begin{enumerate}
\item
Each of qBBs is checked, and if they overlap, the intersection areas are computed and registered as a new qBB; This process is repeated until no new qBBs are found. 
\item
For each $i$-th qBB, the number of overlaps, $N(i)$, is computed and related ranked lists are assigned, and then, the assigned ranked lists are fused by the function $R(i)$.
\end{enumerate}
}{
\begin{equation}
R(i)=N(i) \times \sum_{k=1}^{N(i)} \frac{1}{R_k(i)},
\end{equation}
}

\section{Experiments}\label{sec:exp}

\SW{
To test the performance of the proposed method, we performed extensive experiments on change detection, using the publicly available NCLT (North Campus Long-Term) dataset \cite{nclt}. For each test data, we compared our pixel-wise rank fusion method with score fusion methods, which rely on the availability of relevance scores. Note that this score fusion method can be viewed as an adaptation of previous non-ranking-based (anomaly-based) change detection methods \cite{christiansen2016deepanomaly} to our problem domain. We used it to evaluate the level of achievement of the proposed method. 
}{
}

\SW{
The NCLT dataset is a large-scale, long-term autonomy dataset for robotics research collected at the University of Michigan's North Campus by a Segway vehicle robotic platform. The data we used in the research includes view image sequences along vehicle's trajectories acquired by the front facing
camera of the Ladybug3 with GPS. From the viewpoint of change detection benchmark, the NCLT dataset has desirable properties: 
\begin{enumerate}
\item
It involves both indoor and outdoor change events during seamless indoor and outdoor navigations of the Segway robot.
\item
It contains not only typical changed objects such as cars and pedestrians, but also various kinds of changes such as building construction, construction machines, posters, tables and whiteboards with wheels.
\item
It has been recently widely used in robotics communities as experimental benchmarks for various tasks, such as self-localization \cite{jmangelson-2018a}.
\end{enumerate}
In the current study, we use four datasets 
``2012/1/22", ``2012/3/31", ``2012/8/4", and ``2012/11/17"
(referred to as WI, SP, SU, AU)
collected across different four seasons.
Fig. \ref{fig:env} shows 
the experimental environment
and examples of changed objects
in the dataset.
}{
}

\editage{
\figC
}{}

\editage{
\tabA{}{}
}{}

\SW{
We formulated change detection as a binary classification problem. Each test sample consists of the pairing of two images from two different seasons, referred to as query and reference (or background) images. To create a test set, each image is viewed as a query image candidate and is paired with a reference image with a nearest neighbor viewpoint,
forcing it to belong to a different season
whose viewing angle must be nearer than
45 degree from that of the query image. The test set consists of positive and negative sets. A positive set consists of positive image pair samples, whose query image contains changes with respect to the counterpart reference image.
A negative set consists of those image pairs with no change.
For the positive set, these changed objects are manually annotated in the form of bounding boxes.
The total number of positive and negative samples are 1,054 and 4,188. Image size is 1,232$\times$1,616. Note that successive frames in the NCLT dataset often contain identical and visually similar changed objects in successive frames. To make our test set contain various changed objects, we sampled at most one query image from such successive frames that contain such a visually very similar object. Thus, we obtained 1,054 positive image pairs from the 4$\times$3 season pairs. Fig. \ref{fig:env} shows the robot's viewpoint trajectories and representative changed objects in the test set. 
}{
}

\SW{
For the test system, the visual appearance of each subimage is described by the AE as a feature vector. Feature vectors are indexed and retrieved by a nearest neighbor engine. Prior to the indexing, every feature vector is L2 normalized. For the nearest neighbor, the L2 distance is used as dissimilarity metric. Two independent engines are then employed for the two different object proposal methods (i.e., supervised and unsupervised methods). Most time consuming part of the test system are YOLO-based object proposal and AE-based part feature extraction, which were 11.1 msec per image and 0.9 msec per subimage (Geforce GTX Titan). The number of subimages per image was 15.9 $\pm$ 3.0 (mean$\pm$std), and thus total time cost was 11.1 + 0.9 $\times$ 15.9 = 25.41 msec.
}{
}

\SW{
Note that the length of the rank list can differ among query subimages. Therefore, raw rank values are not comparable among different queries. To address this issue, we normalize the raw rank values by the rank list length to the [0,1]-interval. In our preliminary experiments, we also tested two variants of the proposed system: 
\begin{enumerate}
\item
One where the rank was not normalized, and 
\item
Another where reference subimages were sorted not in descending order but in ascending order of the LoC prior to the rank fusion. 
\end{enumerate}
We found that both variants did not work well and yielded significant performance drops (e.g., AP $<$ 0.6). Moreover, we tested another variant where SIFT's Harris-Laplace region and descriptor were used in place of qBBs and the AE descriptor, finding such a variant also does not work well. The reason might be that such a small Harris-Laplace region could not provide contextual and semantic information that are rich enough for change detection.
}{
}

\SW{
Difficulty indices play an important role in object detection literature standardizing benchmark datasets \cite{radenovic2018revisiting}. In the current study, we introduce two types of difficulty indexes: {\it RoC} and {\it SoB}. {\it RoC} (rate of change) is defined as the area of changed object (i.e., overlap region between the qBB and the ground-truth change qBB) normalized by the area of qBB. If {\it RoC} value is too small for a given qBB, it is difficult to detect such a small changed object. Thus, difficulty of change detection will increase. 
Effectiveness of this overlap-based difficulty index was verified in our previous study on self-localization \cite{kanji2016self}. 
{\it SoB} (size of bounding box) is defined as the area of qBB normalized by the area of image. If the {\it SoB} value is too small for a given qBB, such a subimage looks very dissimilar from the original non-cropped full image,
and the probability of obtaining inappropriate self-localization result will be higher, because typical self-localization systems implicitly or explicitly assume that image size is the same or similar between the input query image and the corresponding reference image. We sampled a test query image set so that query images in the positive and negative sets respectively have restricted {\it RoC} values within their respective $[RoC_{min}^{ + }, RoC_{max}^{ + }]$
and 
$[RoC_{min}^{ - }, RoC_{max}^{ - }]$,
and restricted {\it SoB} values within
their respective
$[SoB_{min}^{ + }, SoB_{max}^{ + }]$
and 
$[SoB_{min}^{ - }, SoB_{max}^{ - }]$.
By default,
we set
$[RoC_{min}^{ + }, RoC_{max}^{ + }]=[0.9, 1]$,
$[RoC_{min}^{ - }, RoC_{max}^{ - }]=[0, 0.05]$,
$[SoB_{min}^{ + }, SoB_{max}^{ + }]=[0, 0.4]$,
and
$[SoB_{min}^{ - }, SoB_{max}^{ - }]=[0.4, 1]$.
}{
$[RoC_{min}^{ - }, RoC_{max}^{ - }]$,
$[SoB_{min}^{ + }, SoB_{max}^{ + }]$
$[SoB_{min}^{ - }, SoB_{max}^{ - }]$
$[RoC_{min}^{ + }, RoC_{max}^{ + }]=[0.4, 1.0]$
$[RoC_{min}^{ - }, RoC_{max}^{ - }]=[0, 0.4]$,
$[SoB_{min}^{ + }, SoB_{max}^{ + }]=[0.9, 1.0]$
$[SoB_{min}^{ - }, SoB_{max}^{ - }]=[0.0, 0.05]$.
}

\SW{
We create various test sets with different levels of difficulty by using {\it RoC} as a difficulty index. More formally, the difficulty of a given test query image set is controlled by two thresholds $RoC_{min}^{ + }$ and $RoC_{max}^{ - }$. If, the {\it RoC} value for a qBB exceeds the threshold of $RoC_{min}^{ + }$ then the qBB is assigned the ground-truth change label of ``change".
If, the {\it RoC} value falls below the threshold of $RoC_{max}^{ - }$ then the qBB is assigned the label of ``no-change". Other qBBs assigned neither ``change" nor ``no-change" are not considered as members of the test set. We created a collection of test sets with different levels of difficulty by changing these two threshold values.
}{
}

\SW{
We compared six different change detection schemes: ``rank fusion", ``rank w/o fusion", ``rank fusion ($\le 2$)", ``rank fusion ($\le 3$)", ``score max", and ``score sum". ``Rank fusion" is the proposed algorithm. ``Rank w/o fusion" is a variant that differs from the proposed algorithm because it does not fuse qBBs to create new qBBs. Therefore, it does not perform pixel-wise rank fusion. ``Rank fusion ($\le k$), $k$= 2, 3" are motivated by the question ``how many pixel-wise rank fusions are required to obtain optimal performance?". Unlike the proposed algorithm, they perform pixel-wise rank fusion at most $k$-times per pixel. If there exists more than $k$ overlapping subimages for a pixel, $k$ subimages are randomly selected and used for that pixel. ``Score max" and ``score sum" are adaptations of ``score max" and ``score sum" strategies in MMR \cite{atrey2010multimodal} per our application domain: the pixel-wise LoC map. These differ from the proposed algorithm, in that relevance scores are used in place of rank values (i.e., assuming the availability of relevance scores output by the self-localization system). For the relevance score, we use L2 distance of feature vectors. The ``score max" fuses pixel-wise relevance score values $v_1$, $\cdots$, $v_k$ by fusion rule 
\begin{equation}
v=\max_{i=1}^k v_i. 
\end{equation}
The ``score sum" fuses them by fusion rule 
\begin{equation}
v=\sum_{i=1}^k v_i. 
\end{equation}
}{
}

\SW{
We investigated 
change detection performance for different $RoC_{max}^{ - }$ values. Generally, if a changed object looks small, its change detection will be more difficult. This experiment is motivated by the question ``how small an object can achieve acceptable change detection performance?".
Table \ref{tab:a} 
shows the AP performance for different $RoC_{max}^{ - }$ values. All the methods tested are not sensitive to the parameter values. Overall, the ``score max" strategy yielded the best performance, whereas it requires the availability of relevance scores output by the self-localization system and relies on the raw-score-merging hypothesis. The proposed ``rank fusion" strategy yielded comparable performance, despite it not requiring relevance scores. It clearly outperformed the ``rank w/o fusion" strategy and other variants.
}{
}

\SW{
}{
}

\SW{
}{
}

\SW{
}{
}

\SW{
In summary, the proposed method successfully detected visually small objects from small (i.e., low-resolution) subimages. Moreover, change detection performance was higher when pixel-wise rank fusion was used. The proposed rank-fusion scheme is comparable to score-fusion schemes, which require the availability of relevance scores. Whereas the current study focuses on combining AE based feature vectors with pixel-wise rank fusion, in a future work, we plan to explore other self-localization schemes, such as different features (e.g., hand-crafted features), and heterogeneous self-localization systems (e.g., particle filters).
}{
}

\section{Conclusions and Future Works}

\SW{
We presented a new detection-by-localization method for object-level change detection using a ranking-based self-localization model, a novel model to evaluate the likelihood-of-change (LoC) of a given query subimage, and unsupervised rank fusion. The ranking-based model did not require training data, nor did it rely on raw-score-merging hypotheses. We consider the proposed unsupervised framework as beneficial to many visual SLAM systems, because such an unsupervised model can generalize it to a broad class of self-localization systems. Future work will investigate ways to use computational resources of different self-localization systems for large scale change detection and map maintenance applications. We will analyze its asymptotical behavior when the number of map databases and self-localization engines increases.
}{
}

\bibliographystyle{IEEEtran}
\bibliography{cd18}

\end{document}